\documentclass{article}
\usepackage{spconf}
\usepackage{amsmath,amssymb,amsfonts}
\usepackage{graphicx}
\usepackage{textcomp}
\usepackage{xcolor}
\usepackage{comment}
\def\BibTeX{{\rm B\kern-.05em{\sc i\kern-.025em b}\kern-.08em
    T\kern-.1667em\lower.7ex\hbox{E}\kern-.125emX}}

\usepackage[colorlinks,urlcolor=blue,linkcolor=blue,citecolor=blue]{hyperref}
\usepackage{color,array}
\usepackage[dvipsnames]{xcolor}
\usepackage{comment}
\usepackage{algorithm}
\usepackage{algpseudocode}
\usepackage{tabu}
\usepackage{amsfonts}
\usepackage{physics}
\usepackage{subfigure}
\usepackage{graphics}
\usepackage{subcaption}
\usepackage{lipsum}

\newcommand{\thickhline}{%
    \noalign {\ifnum 0=`}\fi \hrule height 1pt
    \futurelet \reserved@a \@xhline
}

\title{Time-Series Classification with Multivariate Statistical Dependence Features}
\name{Author(s) Name(s)\thanks{Thanks to XYZ agency for funding.}}
\address{Author Affiliation(s)}

\name{Yao Sun, Bo Hu, Jos\'{e} Pr\'{i}ncipe} 
  
\address{Department of Electrical and Computer Engineering\\ 
University of Florida \\
Gainesville, USA}

\begin{document}
%
\maketitle
\begin{abstract}
In this paper, we propose a novel framework for non-stationary time-series analysis that replaces conventional correlation-based statistics with direct estimation of statistical dependence in the normalized joint density of input and target signals, the cross density ratio (CDR). Unlike windowed correlation estimates, this measure is independent of sample order and robust to regime changes. The method builds on the functional maximal correlation algorithm (FMCA), which constructs a projection space by decomposing the eigenspectrum of the CDR. Multiscale features from this eigenspace are classified using a lightweight single-hidden-layer perceptron. On the TI-46 digit speech corpus, our approach outperforms hidden Markov models (HMMs) and state-of-the-art spiking neural networks, achieving higher accuracy with fewer than 10 layers and a storage footprint under 5 MB.

\end{abstract}
%

\section{Introduction}
\label{sec:intro}
A central challenge in time-series analysis is the accurate estimation of statistics for non-stationary random processes. Conventional methods, such as the Wiener filter \cite{haykin2002adaptive}, estimate autocorrelation and cross-correlation over fixed windows or filter taps. For non-stationary signals, however, such estimates are biased: large windows mix statistics from different regimes, while small windows yield unreliable estimators and require precise segmentation into quasi-stationary intervals. Statistical modeling methods such as hidden Markov models (HMMs) \cite{rabiner} can model regime switching explicitly, but they are complex and require large training datasets.


This paper suggests an alternative to focus on the joint probability density function (PDF) of input and target signals, which is independent of sample order and thus more robust to regime changes. This approach captures the statistical dependence structure, a key element in many machine learning and signal processing tasks, including feature selection \cite{brown2012conditional}, dimensionality reduction \cite{bach2002kernel}, causal inference \cite{shimizu2006linear}, and predictive modeling \cite{ishmael2018mine}. Yet, conventional dependence measures remain limited: Pearson correlation \cite{pearson} captures only linear relationships, while mutual information \cite{information} quantifies full dependence but reduces it to a scalar, lacks specificity in high-dimensional distribution estimations. Adaptive filters (e.g., LMS \cite{haykin2002adaptive}) and recurrent neural networks \cite{medsker2001recurrent} improve discrimination between regimes but do not fundamentally solve the statistical mixing problem.


The functional maximal correlation algorithm (FMCA) \cite{fmca} provides a different perspective. Instead of relying on temporal correlations, FMCA estimates the joint PDF of input and target signals, allowing stable density estimation from long or randomized windows of non-stationary data. From this, FMCA constructs an eigenspace that captures rich multivariate dependencies, yielding principled feature representations for prediction and classification while avoiding the tradeoffs of conventional windowing methods.

Building on this foundation, we propose an FMCA-based framework for non-stationary time-series classification. The framework comprises two neural networks trained with a correlation-maximizing objective, a feature aggregation module that computes power-based features across multiple time scales, and a lightweight classifier for label assignment. We evaluate this framework on speech-based word recognition, a challenging task due to the inherent non-stationarity of speech. Results show that FMCA-derived features are robust and efficient, achieving competitive recognition accuracy at low computational cost.

The primary contributions of this work are as follows: First, we apply the FMCA methodology to nonstationary time series classification. Second, we conduct a comprehensive comparison between temporal and spectral input representations, demonstrating that spectral features enhance classification accuracy by ordering the input by frequency components. Finally, we design a compact FMCA-based framework that achieves robust time-series recognition using lightweight network structures, demonstrating excellent scalability.

\section{Methods}
\label{sec:method}

\subsection{Construct a Projection Space to Measure Statistical Dependence with FMCA}
The goal of the functional maximal correlation algorithm (FMCA) is to construct a multivariate feature space that captures complex dependencies between two random processes, \(\mathbf{x} = \{\mathbf{x}(t), t \in \mathcal{T}_1\} \) and \(\mathbf{u} = \{\mathbf{u}(t), t \in \mathcal{T}_2\},\) with joint density $p(x,u)$ and marginals $p(x)$ and $p(u)$. FMCA operates by performing an orthonormal spectral decomposition of the cross density ratio (CDR) $\rho(x,u)$:
\begin{equation}
\begin{gathered}
\rho(x, u)= \frac{p(x, u)}{p(x)p(u)} = \sum_{k=1}^{\infty} \lambda_k \hat{\phi}_k(x) \hat{\psi}_k(u) , \\
	\int_{\mathcal{X}} \hat{\phi}_i(x) \hat{\phi}_j(x) p(x) dx = \delta(i,j), \\
	\int_{\mathcal{U}} \hat{\psi}_i(u) \hat{\psi}_j(u) p(u) du = \delta(i,j),
\end{gathered}
\end{equation}
where $\delta(i,j)$ represents a delta function (1 when $i=j$, 0 otherwise). The CDR is positive-definite and thus defines a reproducing kernel Hilbert space (RKHS) based on the data distribution \cite{scholkopf2002learning}. According to Mercer's theorem \cite{rkhs}, $\rho(x,u)$ can be represented as a spectral decomposition with eigenvalues $\{\lambda_k\}_{k=1}^{\infty}$, and eigenfunctions $\{\hat{\phi}_k\}_{k=1}^{\infty}$ and $\{\hat{\psi}_k\}_{k=1}^{\infty}$.

\begin{figure}[t]
	\centering
	\includegraphics[width=.45\textwidth]{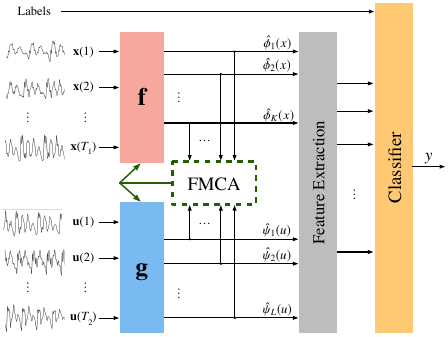}
	\caption{Overview of the proposed FMCA-based framework for non-stationary time-series classification.}
    \vspace{-.8em}
	\label{fmca_networks}
\end{figure}

To approximate this decomposition in practice, FMCA employs two neural networks, $\mathbf{f}_{\theta} : \mathcal{X} \rightarrow \mathbb{R}^K$ and $\mathbf{g}_{\omega} : \mathcal{U} \rightarrow \mathbb{R}^K$, that project inputs $\mathbf{x}$ and $\mathbf{u}$ into a $K$-dimensional output space. As shown in Fig.~\ref{fmca_networks}, the networks are optimized to minimize the following cost function:
\begin{equation}
\label{eqn:cost}
\begin{gathered}
r^{\ast} = \min_{\theta, \omega} r(\mathbf{f}_{\theta}, \mathbf{g}_{\omega}),\\
r(\mathbf{f}_{\theta}, \mathbf{g}_{\omega}) = \log \det \mathbf{R}_{FG} - \log \det \mathbf{R}_F - \log \det \mathbf{R}_G, 
\end{gathered}
\end{equation}
where $\mathbf{R}_F$ and $\mathbf{R}_G$ denote the autocorrelation functions (ACFs) of the network outputs, and $\mathbf{P}_{FG}$ denotes their cross-correlation function (CCF):
\begin{equation}
\begin{gathered}
\mathbf{R}_F = \mathbb{E}_\mathbf{x}[\mathbf{f}_{\theta}(\mathbf{x})\mathbf{f}_{\theta}^\intercal(\mathbf{x})], \mathbf{R}_G = \mathbb{E}_\mathbf{u}[\mathbf{g}_{\omega}(\mathbf{u})\mathbf{g}_{\omega}^\intercal(\mathbf{u})], \\
\mathbf{P}_{FG} = \mathbb{E}_{\mathbf{x}, \mathbf{u}}[\mathbf{f}_{\theta}(\mathbf{x})\mathbf{g}_{\omega}^\intercal(\mathbf{u})], 
\mathbf{R}_{FG} = \begin{bmatrix}
\mathbf{R}_F & \mathbf{P}_{FG} \\
\mathbf{P}_{FG}^\intercal & \mathbf{R}_G 
\end{bmatrix}.
\end{gathered}
\end{equation}
A small diagonal matrix $\epsilon \mathbf{I}$ is added to ACFs for regularization purpose.

Once the networks converges, their outputs are normalized:
\begin{equation}
\overline{\mathbf{f}_{\theta}} = \mathbf{R}_F^{-\frac{1}{2}} \mathbf{f}_{\theta}, 
\quad \overline{\mathbf{g}_{\omega}} = \mathbf{R}_G^{-\frac{1}{2}} \mathbf{g}_{\omega},
\end{equation}
ensuring orthonormality of the transformed functions without affecting the cost. Finally, the eigen-expansion can be computed with singular value decomposition (SVD):
\begin{equation}
\begin{aligned}
    \overline{\mathbf{P}}_{FG} = \mathbb{E}\left[\overline{\mathbf{f}_{\theta}}(\mathbf{x}) \overline{\mathbf{g}_{\omega}} ^{\intercal}(\mathbf{u})\right], 
    \quad  \overline{\mathbf{P}}_{FG} \overline{\mathbf{P}}_{FG}^{\intercal} = \mathbf{Q}_F \mathbf{\Sigma} \mathbf{Q}_F^{\intercal}, \\
    \overline{\mathbf{P}}_{FG}^{\intercal} \overline{\mathbf{P}}_{FG} = \mathbf{Q}_G \mathbf{\Sigma} \mathbf{Q}_G^{\intercal}, 
    \quad  \mathbf{\Sigma} = \begin{bmatrix}
        \sigma_1 & & \\
        & \ddots & \\
        & & \sigma_K
    \end{bmatrix},
\end{aligned}
\end{equation}
where $\{\sigma_i\}_{i=1}^K$ contains the top $K$ eigenvalues of the CDR. The normalized eigenfunctions are then obtained as: 
\begin{equation}
\hat{\mathbf{f}}_{\theta} = \mathbf{Q}_F^{\intercal} \overline{\mathbf{f}_{\theta}}, \quad \hat{\mathbf{g}}_{\omega} = \mathbf{Q}_G^{\intercal} \overline{\mathbf{g}_{\omega}}. 
\end{equation}

By minimizing the cost, FMCA finds the leading eigenfunctions, providing an accurate finite-dimensional approximation of the CDR, where $K$ is a user-defined hyperparameter.

\begin{figure*}[t]
     \centering
     \subfigure[Eigenfunction projections over the projective space built based on framed time-domain speech segments. Some digit classes, such as “two” and “three,” exhibit highly similar projection patterns, leading to potential confusion.]{
         \centering
         \label{eig_no_fft}
         \vspace{-.5em}
         \includegraphics[width=0.46\textwidth]{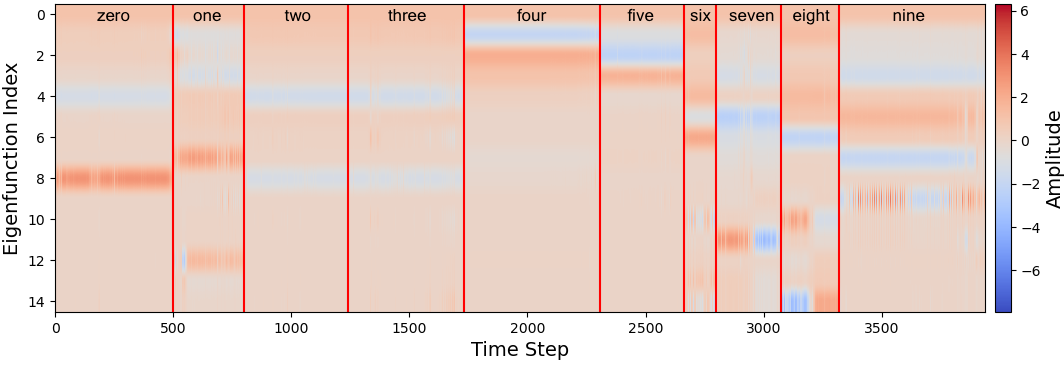}}
     \subfigure[Eigenfunction projections obtained from frequency-domain input representations. Compared to temporal inputs, spectral features produce clearer separability between digits and tighter intra-class clustering.]{
         \centering
         \label{eig_fft}
         \includegraphics[width=0.46\textwidth]{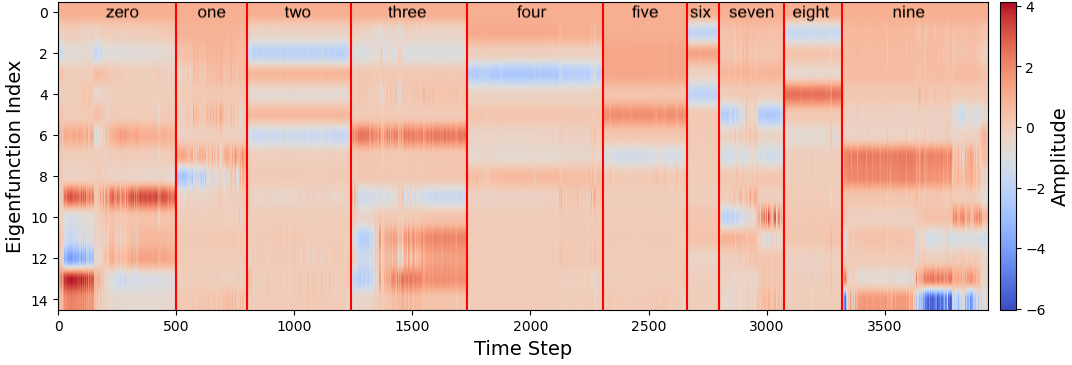}}
     \vspace{-.8em}
     \caption{Eigenfunction projections across time for digits ``zero" through ``nine'' using two input representations.}
     \vspace{-.5em}
     \label{no_fft_result}
\end{figure*}

\subsection{Feature Extraction and Classification}

For classification, the frames from the time-series signals are used to learn the projection space of the joint distribution. After training, the input data network yields an orthonormal basis $\{\hat{\phi}_i(x)\}$ that defines the projection space. Once the FMCA networks converge, each input signal is projected onto this space to obtain time-varying eigenfunction coefficients, forming a sequence of projection vectors.

A feature matrix is then constructed by computing the power of these projections across multiple time intervals, effectively capturing dynamic variations in the signal.  These power-based features represent the eigenfunction power density, and serves as inputs to a lightweight classifier. In this work, we employ a single-hidden-layer multilayer perceptron (MLP) to perform signal classification, although the framework accommodates any type of classifiers.

\section{Experiments}
\label{sec:experiments}
In this section we evaluate the proposed FMCA framework on the TI-46 isolated digits dataset \cite{ti46}, which contains 4,000 utterances of digits “zero”-“nine” from eight female and eight male speakers (400 recordings per digit). Speech is inherently non-stationary, with rapid spectral and temporal changes due to phoneme transitions, coarticulation, and speaker variability \cite{rabiner1993fundamentals, deng2003speech}, making TI-46 both a challenging and suitable testbed for dependence-based modeling.

\subsection{Data Preprocessing}
\label{sec:preprocess}

Each speech signal undergoes three preprocessing steps: 1. Normalized to range [-1, 1]. 2. Truncated to remove silence using a threshold-based endpoint detector. 3. Segmented into $W$ frames of $L$ samples with stride $S$. No zero-padding is applied, and $L, S$ are fixed across train/test sets. 

\subsection{System Setup}

FMCA networks $\mathbf{f}, \mathbf{g}$ share identical architectures, with input size $L$ (temporal) or $L/2$ (spectral), and $K$ outputs corresponding to leading CDR eigenfunctions. Each has $n_1$ fully connected layers with $H_1$ units and layer normalization. A small constant $\epsilon$ regularizes updates.

During training, $\mathbf{x}, \mathbf{u}$ are randomly drawn from the same class but not necessarily the same speaker, ensuring speaker independent experiments. The final FMCA layer applies a softmax activation to produce eigenfunction projections over time. For each signal, a $K$-dimensional feature vector is obtained by computing the power of the projections. To capture temporal variations, features are computed across $T$ evenly divided temporal intervals within each utterance, resulting in a $K \times T$ feature matrix that embeds time-localized statistical dependencies. These matrices are then flattened into 1D vectors and classified using a single-hidden-layer MLP with $H_2$ hidden units. Both FMCA networks and classifier are optimized with Adam \cite{adam}, with separate learning rates $lr_1$ and $lr_2$. During inference only the projections from $\mathbf{f}$ is used.

\subsection{Time vs. Frequency-Domain Representations}
\label{sec:time-fre}


We compare two approaches of input representations: utilizing either time or frequency-domain information. The temporal approach uses waveform segments. Alternatively, the spectral approach uses the magnitude spectra obtained via a fast Fourier transform (FFT), reducing input window length to $L/2$ by discarding redundant data.

Fig. \ref{no_fft_result} compares eigenfunction projections across time for randomly picked training utterances of digits “zero” through “nine” under both representations. In the temporal setting (Fig. \ref{eig_no_fft}), projections for “two” and “three” show highly similar patterns, causing significant confusion between these classes. By contrast, frequency-domain representations (Fig. \ref{eig_fft}) produce more distinct projections, resulting in tighter intra-class clustering and greater inter-class separability.

This improvement arises because the FFT reorders inputs by frequency, simplifying the learning of temporal patterns that may occur anywhere within a speech segment. Consequently, all subsequent experiments use spectral features.

\subsection{Influence of Model Parameters}
\label{sec:para}

We examine four key hyperparameters: window size $L$, stride $S$, projection space dimension $K$, and feature extraction intervals $T$. All reported accuracies are averaged over 20-fold cross-validation, using a 4:1 train/test split balanced across speakers and classes.

\begin{figure}[t]
	\vspace{-.7em}
	\centering
	\includegraphics[width=.4\textwidth]{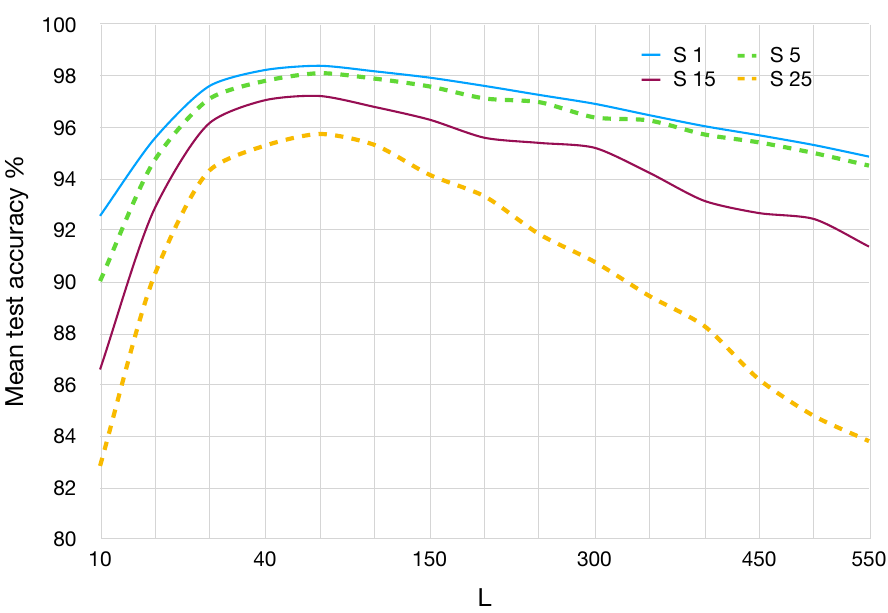}
	\vspace{-.6em}
	\caption{Impact of window size $L$ and stride $S$ on classification accuracy.}
	\label{r_l_s}
\end{figure}

\begin{table}[t]
\centering
\caption{Fixed parameters for experiments in Fig.~\ref{r_l_s}.}
\vspace{-.5em}
\label{conf_1}
\resizebox{0.9\linewidth}{!}{
\begin{tabular}{ccccccccc}
	\hline
	$H_1$ & $n_1$ & $lr_1$ & $\epsilon$ & $I$ & $K$ & $T$  & $H_2$ & $lr_2$  \\
	\hline
	200 & 3 & 0.001 & 1e-4 & 7000 & 8 & 6 & 40 & 0.001 \\
	\hline
    \vspace{-.8em}
\end{tabular}}
\end{table}

With other parameters fixed as in Table~\ref{conf_1}, Fig. \ref{r_l_s} shows that smaller strides consistently improve accuracy, with the best results achieved at $S$ = 1, where projections are computed for every sample point, fully leveraging temporal resolution. An optimal window size of $L \approx 50$ yields the highest accuracy. Shorter windows ($L < 30$) fail to capture sufficient frequency information, while overly large windows mix multiple phonemes within a frame, degrading discriminability especially for shorter words like “six” and “eight.”

\begin{figure}[t]
\centering
\includegraphics[width=.4\textwidth]{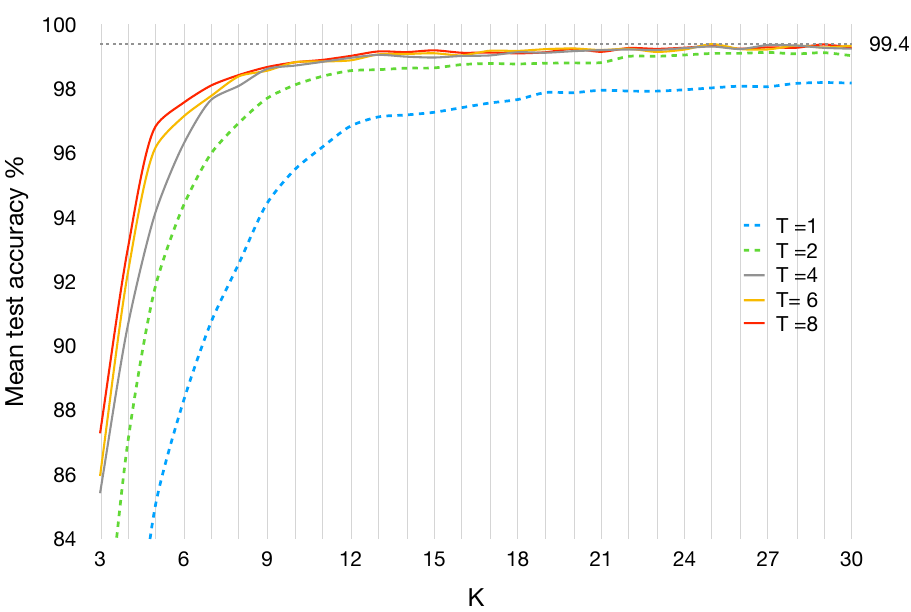}
\vspace{-.7em}
\caption{Effects of projection space dimension $K$ and feature extraction times $T$.}
\vspace{-.8em}
\label{r_k_t}
\end{figure}

Fig. \ref{r_k_t} show that increasing $K$ improves accuracy up to approximately $K$ = 20, beyond which gains plateau. Similarly, increasing $T$ enhances performance until $T$ = 4, after which additional temporal subdivisions provide negligible benefits. Here, $T$ serves as the temporal evaluation density. Interestingly, the optimal $T$ aligns with the phonetic structure of the dataset that most digits except “seven” contain four or fewer phonemes, suggesting that four intervals are sufficient to capture relevant temporal variations. 

Larger values of $K$ and $T$ also produce feature matrices with higher dimensions, which require a larger hidden dimension $H_2$ in the classifier for effective classification. Furthermore, because utterances vary in length, the samples spanned by each of the $T$ intervals differ across signals. This means the MLP classifier operates on signal-specific temporal embeddings rather than the fixed embedding sizes typical of conventional adaptive filters. Finally, the tuning of all these hyperparameters above depends on the statistical characteristics of the experimental dataset. When applying the framework to a new dataset, they should be re-adapted accordingly.

\subsection{Comparison with Other Models}

In this section, we compare our FMCA-based system with several reported resource-efficient speech recognition methods. For a fair comparison, we focus on models that are trained and tested exclusively on TI-46 dataset, excluding more complex systems pre-trained on extensive datasets. Table~\ref{table:compare} summarizes the results.

\textbf{HMM Baseline:} Each utterance is pre-emphasized with $\alpha=0.95$, framed into 400-sample windows with 240-sample overlap, and smoothed using a Hamming window. Thirteen MFCCs \cite{mfcc} are extracted per frame and fed to a 5-state HMM isolated word recognizer. Performance is evaluated with 5-fold cross-validation (4:1 split), yielding a mean accuracy of 96.48\%.

\textbf{Spike-based Systems:} Spike-train KAARMA \cite{spike} attains 95.23\% accuracy on the full dataset using a 2:1 train/test split.
Digital LSM \cite{lsm}, a liquid state machine variant, reports 92.30\% accuracy on a 1,590-utterance subset and 99.79\% accuracy on a much smaller 500-utterance subset fine-tuned for five speakers. However, this near-perfect result comes at the expense of generalizability.
SWAT-SNN \cite{swat} achieves 95.25\% on a 400-sample subset (8 speakers $\times$ 10 digits $\times$ 5 utterances).

\textbf{Proposed FMCA-based model:} Our model achieves 99.39\% on the full dataset, 98.96\% on the 500-utterance subset, and 97.78\% on the 400-utterance subset. Unlike Digital LSM, which heavily depends on speaker-specific tuning, our method maintains high accuracy across speakers and subsets without requiring dataset-specific optimization.

\begin{table}[t]
\centering
\tabcolsep=0.01cm
\caption{Comparisons with other reported methods.}
\vspace{-.6em}
\label{table:compare}
\resizebox{0.95\linewidth}{!}{
	\begin{tabular}{cccccccccc}
		\hline
			&	$\#$ 	& $\#$ &	Train/Test 	&	Test  \\
        & Speakers &  Samples & Ratio & Accuracy \\
		\hline
		5-state HMM	&	16	&	4000 & 4:1	&	96.48$\%$  \\	
        \hline
		Spike-train KAARMA 	&	16	&	4000 & 2:1	&	95.23$\%$ 	\\
        \hline
 
		Digital LSM 	&	16	&	1590 & 4:1	&	92.30$\%$  \\ 
		  &	5	&	500	& 4:1 & $\mathbf{99.79\%}$ 	\\
        \hline
    
		SWAT SNN	&	8	&	400 & 4:1	&	95.25$\%$ 	\\
        \hline

		FMCA	&	16	&	4000  & 4:1	&	$\mathbf{99.39\%}$ 	\\
				&	5	& 500	&  4:1	&	98.96$\%$  \\
				&	8	& 400	&  4:1	&	$\mathbf{97.78\%}$ \\
		\hline
	\end{tabular}}
    \vspace{-.8em}
\end{table}

\textbf{Efficiency and Scalability:} Spike-based systems generally rely on memory-intensive, recurrent architectures, limiting scalability and increasing computational cost. In contrast, our FMCA-based system uses a simple, feedforward structure, training within 10 minutes on a single NVIDIA RTX A5000 GPU, which is significantly shorter compared to 5.16 hours for Digital LSM.

\section{Conclusion}
\label{sec:conclusion}

We propose a novel FMCA-based framework for time-series classification that constructs a Hilbert space representation from the probability density functions of input signals. By focusing on PDF estimation rather than windowed temporal correlation measures, the system avoids the statistical mixing problem across non-stationary regimes and extracts high-quality features using lightweight neural networks. Experiments on the TI-46 dataset demonstrate that our approach outperforms several state-of-the-art compact models while maintaining low computational cost, making it well-suited for on-device applications. As future work, we plan to extend the framework to the complex-valued domain for improved spectral information processing.

\clearpage

\bibliographystyle{IEEEbib}
\bibliography{refs.bib}

@article{fmca,
  title={The cross density kernel function: A novel framework to quantify statistical dependence for random processes},
  author={Hu, Bo and Principe, Jose C},
  journal={arXiv preprint arXiv:2212.04631},
  year={2022}
}

@article{adam,
  title={Adam: A method for stochastic optimization},
  author={Kingma, Diederik P and Ba, Jimmy},
  journal={arXiv preprint arXiv:1412.6980},
  year={2014}
}

@article{rabiner,
  title={A tutorial on hidden Markov models and selected applications in speech recognition},
  author={Rabiner, Lawrence R},
  journal={Proceedings of the IEEE},
  volume={77},
  number={2},
  pages={257--286},
  year={1989},
  publisher={Ieee}
}

@article{spike,
  title={Biologically-inspired spike-based automatic speech recognition of isolated digits over a reproducing kernel Hilbert space},
  author={Li, Kan and Principe, Jose C},
  journal={Frontiers in neuroscience},
  volume={12},
  pages={275461},
  year={2018},
  publisher={Frontiers}
}

@article{mfcc,
  title={Comparison of parametric representations for monosyllabic word recognition in continuously spoken sentences},
  author={Davis, Steven and Mermelstein, Paul},
  journal={IEEE transactions on acoustics, speech, and signal processing},
  volume={28},
  number={4},
  pages={357--366},
  year={1980},
  publisher={IEEE}
}

@article{rkhs,
  title={Theory of reproducing kernels},
  author={Aronszajn, Nachman},
  journal={Transactions of the American mathematical society},
  volume={68},
  number={3},
  pages={337--404},
  year={1950}
}

@misc{ti46,
  author = {Mark Liberman et al},
  title = {TI 46-Word},
  year = {1993},
  note = {Philadelphia: Linguistic Data Consortium, 
          \url{https://doi.org/10.35111/zx7a-fw03}}
}

@article{lsm,
  title={A digital liquid state machine with biologically inspired learning and its application to speech recognition},
  author={Zhang, Yong and Li, Peng and Jin, Yingyezhe and Choe, Yoonsuck},
  journal={IEEE transactions on neural networks and learning systems},
  volume={26},
  number={11},
  pages={2635--2649},
  year={2015},
  publisher={IEEE}
}

@article{swat,
  title={SWAT: A spiking neural network training algorithm for classification problems},
  author={Wade, John J and McDaid, Liam J and Santos, Jose A and Sayers, Heather M},
  journal={IEEE Transactions on neural networks},
  volume={21},
  number={11},
  pages={1817--1830},
  year={2010},
  publisher={IEEE}
}

@incollection{pearson,
  title={Pearson correlation coefficient},
  author={Benesty, Jacob and Chen, Jingdong and Huang, Yiteng and Cohen, Israel},
  booktitle={Noise reduction in speech processing},
  pages={1--4},
  year={2009},
  publisher={Springer}
}

@book{information,
  title={Information theoretic learning: Renyi's entropy and kernel perspectives},
  author={Principe, Jose C},
  year={2010},
  publisher={Springer Science \& Business Media}
}

@book{scholkopf2002learning,
  title={Learning with kernels: support vector machines, regularization, optimization, and beyond},
  author={Sch{\"o}lkopf, Bernhard and Smola, Alexander J},
  year={2002},
  publisher={MIT press}
}

@book{rabiner1993fundamentals,
  title={Fundamentals of speech recognition},
  author={Rabiner, Lawrence and Juang, Biing-Hwang},
  year={1993},
  publisher={Prentice-Hall, Inc.}
}

@book{deng2003speech,
  title={Speech processing: a dynamic and optimization-oriented approach},
  author={Deng, Li and O'Shaughnessy, Douglas},
  year={2003},
  publisher={CRC Press}
}

@article{brown2012conditional,
  title={Conditional likelihood maximisation: a unifying framework for information theoretic feature selection},
  author={Brown, Gavin and Pocock, Adam and Zhao, Ming-Jie and Luj{\'a}n, Mikel},
  journal={The journal of machine learning research},
  volume={13},
  number={1},
  pages={27--66},
  year={2012},
  publisher={JMLR. org}
}

@article{bach2002kernel,
  title={Kernel independent component analysis},
  author={Bach, Francis R and Jordan, Michael I},
  journal={Journal of machine learning research},
  volume={3},
  number={Jul},
  pages={1--48},
  year={2002}
}

@article{shimizu2006linear,
  title={A linear non-Gaussian acyclic model for causal discovery.},
  author={Shimizu, Shohei and Hoyer, Patrik O and Hyv{\"a}rinen, Aapo and Kerminen, Antti and Jordan, Michael},
  journal={Journal of Machine Learning Research},
  volume={7},
  number={10},
  year={2006}
}

@article{ishmael2018mine,
  title={MINE: mutual information neural estimation},
  author={Ishmael Belghazi, Mohamed and Baratin, Aristide and Rajeswar, Sai and Ozair, Sherjil and Bengio, Yoshua and Courville, Aaron and Devon Hjelm, R},
  journal={arXiv e-prints},
  pages={arXiv--1801},
  year={2018}
}

@book{haykin2002adaptive,
  title={Adaptive filter theory},
  author={Haykin, Simon S},
  year={2002},
  publisher={Pearson Education India}
}

@article{medsker2001recurrent,
  title={Recurrent neural networks},
  author={Medsker, Larry R and Jain, Lakhmi and others},
  journal={Design and applications},
  volume={5},
  number={64-67},
  pages={2},
  year={2001}
}

\end{document}